\pdfoutput=1

\documentclass[11pt]{article}

\usepackage{ACL2023}

\usepackage{times}
\usepackage{latexsym}

\usepackage[T1]{fontenc}

\usepackage[utf8]{inputenc}

\usepackage{microtype}

\usepackage{inconsolata}

\usepackage{paralist}
\usepackage{adjustbox}

\usepackage{amsmath,amssymb,amsfonts}
\usepackage{pifont}
\usepackage{utfsym}

\usepackage{multirow}
\usepackage{booktabs}
\usepackage{arydshln}
\usepackage{bm}
\usepackage{color,xcolor}
\usepackage{colortbl}

\newcommand{\tabincell}[2]{\begin{tabular}{@{}#1@{}}#2\end{tabular}}

\definecolor{highlightgreen1}{RGB}{229,250,245}
\definecolor{highlightgreen2}{RGB}{0,120,87}
\definecolor{purple1}{RGB}{175,20,241}
\definecolor{green1}{RGB}{0,176,80}

\definecolor{ired}{RGB}{196,44,46}
\definecolor{igreen}{RGB}{5,209,75}

\definecolor{nmgray}{RGB}{229,229,229}

\title{\emph{Scene Graph as Pivoting}: Inference-time Image-free Unsupervised Multimodal Machine Translation with Visual Scene Hallucination}

\author{
Hao Fei\textsuperscript{\rm 1},  \, 
Qian Liu\textsuperscript{\rm 2}, \, 
Meishan Zhang\textsuperscript{\rm 3}\Thanks{ Corresponding author},  \, 
Min Zhang\textsuperscript{\rm 3},  \, 
Tat-Seng Chua\textsuperscript{\rm 1} \\
\textsuperscript{\rm 1} Sea-NExT Joint Lab, School of Computing, National University of Singapore \\
\textsuperscript{\rm 2} Nanyang Technological University \,
\textsuperscript{\rm 3} Harbin Institute of Technology (Shenzhen) \\
\tt {\{haofei37, dcscts\}@nus.edu.sg \quad  liu.qian@ntu.edu.sg } \\ 
\tt {mason.zms@gmail.com  \quad  zhangmin2021@hit.edu.cn}
}

\begin{document}
\maketitle

\begin{abstract}
In this work, we investigate a more realistic unsupervised multimodal machine translation (UMMT) setup, \emph{inference-time image-free} UMMT, where the model is trained with source-text image pairs, and tested with only source-text inputs.
First, we represent the input images and texts with the visual and language scene graphs (SG), where such fine-grained vision-language features ensure a holistic understanding of the semantics.
To enable pure-text input during inference, we devise a visual scene hallucination mechanism that dynamically generates pseudo visual SG from the given textual SG.
Several SG-pivoting based learning objectives are introduced for unsupervised translation training. 
On the benchmark Multi30K data, our SG-based method outperforms the best-performing baseline by significant BLEU scores on the task and setup, helping yield translations with better completeness, relevance and fluency without relying on paired images.
Further in-depth analyses reveal how our model advances in the task setting.
\end{abstract}

\section{Introduction}

\vspace{-1mm}
Current neural machine translation (NMT) has achieved great triumph \cite{SutskeverVL14,BahdanauCB14,ZhuXWHQZLL20}, however in the cost of creating large-scale parallel sentences, which obstructs the development of NMT for the minor languages.
Unsupervised NMT (UMT) has thus been proposed to relieve the reliance of parallel corpora \cite{ArtetxeLAC18,ChenLL18}.
The core idea of UMT is to align the representation spaces between two languages with alternative pivot signals rather than parallel sentences, such as bilingual lexicons \cite{LampleCDR18}, multilingual language models (LM) \cite{ConneauL19} and back-translation technique \cite{sennrich-etal-2016-improving}.
Recent trends have considered the incorporation of visual information, i.e., multimodal machine translation (MMT) \cite{specia-etal-2016-shared,huang-etal-2016-attention}.
Intuitively, visual modality can serve as language-agnostic signals, pivoting different languages by grounding the same textual semantics into the common visual space.
Therefore, solving UMT with visual contents as pivot becomes a promising solution, a.k.a., unsupervised MMT (UMMT) \cite{huang-etal-2020,SuFBKH19}.

\begin{table}
 \setlength{\tabcolsep}{1.0mm}
\centering
\resizebox{0.98\columnwidth}{!}{
\begin{tabular}{llcc}
\hline
& & \textbf{\tabincell{c}{Avoid parallel sent.\\ during training?}} & \textbf{\tabincell{c}{Avoid paired img.\\ during testing?}} \\
\hline
\rowcolor{nmgray} \multicolumn{4}{l}{$\bullet$ \emph{Supervised MMT}}\\
& General MMT &	\textcolor{ired}{\small \usym{2613}} & \textcolor{ired}{\small \usym{2613}} \\

\cdashline{2-4}
&	\multirow{3}{*}{\tabincell{l}{
\citet{0001C0USLZ20} \\
\citet{fang-feng-2022-neural} \\
\citet{LiPKCFCV22}}} & \multirow{3}{*}{\textcolor{ired}{\small \usym{2613}}} &	\multirow{3}{*}{\textcolor{igreen}{\large \checkmark}}\\ \\ \\

\hline
\rowcolor{nmgray} \multicolumn{4}{l}{$\bullet$ \emph{Unsupervised MMT}}\\
& \multirow{3}{*}{\tabincell{l}{
\citet{ChenLL18}\\
\citet{SuFBKH19} \\
\citet{huang-etal-2020} }} &	\multirow{3}{*}{\textcolor{igreen}{\large \checkmark}} & \multirow{3}{*}{\textcolor{ired}{\small \usym{2613}}} \\ \\ \\

\cdashline{2-4}
& \bf This work &	\textcolor{igreen}{\large \checkmark} &	\textcolor{igreen}{\large \checkmark}  \\
\hline
\end{tabular}
}
\vspace{-1mm}
\caption{
Practical unsupervised MMT requires the avoidance of not only parallel sentences during training, but also the paired image during inference (testing).
}
\vspace{-3mm}
\label{tab:intro}
\end{table}

UMMT systems are trained with only the text-image pairs (<\emph{text-img}>), which can be easier to collect than the parallel source-target sentence pairs (<\emph{src-tgt}>) \cite{huang-etal-2020}.
Although exempting the parallel sentences for training, 
UMMT still requires such text-image pairs as inputs for testing.
Yet such assumption might be unrealistic, because in most of the real-world scenarios such as online translation systems, paired images are not available during inference.
Especially for some scarce languages, the <\emph{text-img}> pairs have difficult access.
In other words, practical UMMT systems should not only avoid \emph{the parallel sentences during training}, but also \emph{the text-image pairs during inference}. 
As summarized in Table \ref{tab:intro}, although some existing MMT researches exempt the testing-time visual inputs \cite{0001C0USLZ20,LiPKCFCV22}, they all unfortunately are supervised methods, relying on large-scale parallel sentences for training.

As emphasized above, the visual information is vital to UMMT.
However, for both the existing supervised and unsupervised MMT studies, they may suffer from ineffective and insufficient modeling of visual pivot features.
For example, most of MMT models perform vision-language (VL) grounding over the whole image and text \cite{huang-etal-2019-multi,0001C0USLZ20}, where such coarse-grained representation learning can cause mismatching and sacrifice the subtle VL semantics.
\citet{fang-feng-2022-neural} recently introduce a fine-grained VL alignment learning via phrase-level grounding, while without a holistic understanding of the visual scene, such local-level method may lead to incomplete or missing alignments.

\begin{figure}[!t]
\centering
\includegraphics[width=0.98\columnwidth]{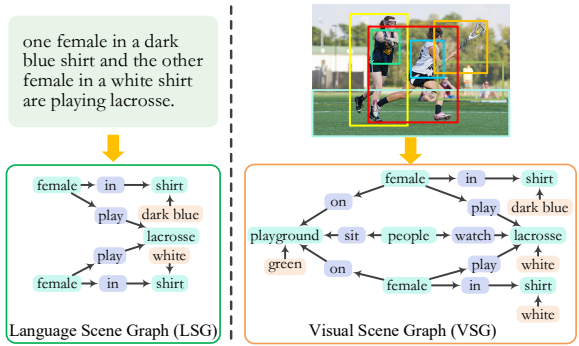}
\vspace{-1mm}
\caption{
Representing the texts and images via language scene graphs (LSG) and visual scene graphs (VSG).
In a scene graph, \emph{object}, \emph{attribute}, \emph{relation} nodes are shown in green, orange, purple respectively. 
}
\label{intro}
\vspace{-3mm}
\end{figure}

\begin{figure}[!t]
\centering
\includegraphics[width=0.98\columnwidth]{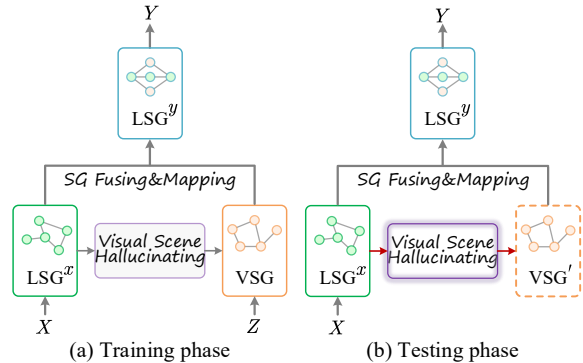}
\vspace{-1mm}
\caption{
The high-level overview of our SG-based UMMT model.
During training, src-side sentences with paired images are used as inputs, together with the corresponding LSG and VSG.
Testing phase only takes src-side sentences, where the visual hallucination module is activated to generate VSG from text sources.
}
\label{framework}
\vspace{-3mm}
\end{figure}

In this work, we present a novel UMMT method that solves all aforementioned challenges.
First of all, to better represent the visual (also the textual) inputs, we consider incorporating the visual scene graph (VSG) \cite{JohnsonKSLSBL15} and language scene graph (LSG) \cite{wang-etal-2018-scene}.
The scene graphs (SG) advance in intrinsically depicting the semantic structures of texts or images with rich details (cf. Fig. \ref{intro}), which offers a holistic viewpoint for more effective pivoting learning.
Then, we build the UMMT framework as illustrated in Fig. \ref{framework}.
The input src text and paired image are first transformed into LSG and VSG, which are further fused into a mixed SG, and then translated into the tgt-side LSG.
And the tgt sentence will be finally produced conditioned on the tgt LSG.
Several SG-based pivoting learning strategies are proposed for unsupervised training of UMMT system.
In addition, to support pure-text (image-free) input during inference, we devise a novel visual scene hallucination module, which dynamically generates a hallucinated VSG from the LSG compensatively.

Our system is evaluated on the standard MMT \emph{Multi30K} and NMT \emph{WMT} data.
Extensive experimental results verify that the proposed method outperforms strong baselines on unsupervised multimodal translation by above 5 BLEU score on average.
We further reveal the efficacy of the visual scene hallucination mechanism in relieving the reliance on image inputs during inference.
Our SG-pivoting based UMMT helps yield translations with higher completeness, relevance and fluency, and especially obtains improvements on the longer sentences.

Overall, we make the following contributions: 

$\blacktriangleright$ 1) We are the first to study the \emph{inference-time image-free} unsupervised multimodal machine translation, solved with a novel visual scene hallucination mechanism.
$\blacktriangleright$ 2) We leverage the SGs to better represent the visual and language inputs.
Moreover, we design SG-based graph pivoting learning strategies for UMMT training.
$\blacktriangleright$ 3) Our model achieves huge boosts over strong baselines
on benchmark data.
Code is available at \url{https://github.com/scofield7419/UMMT-VSH}.

\section{Related Work}

\vspace{-1mm}
Neural machine translation has achieved notable development in the era of deep learning \cite{SutskeverVL14,BahdanauCB14,luong-etal-2015-effective}.
The constructions of powerful neural models and training paradigms as well as the collection of large-scale parallel corpora are the driving forces to NMT's success \cite{VaswaniSPUJGKP17,devlin-etal-2019-bert}.
The key of NMT is to learn a good mapping between two (or more) languages.
In recent years, visual information has been introduced for stronger NMT (i.e., multimodal machine translation), by enhancing the alignments of language latent spaces with visual grounding \cite{specia-etal-2016-shared,huang-etal-2016-attention}.
Intuitively, people speaking different languages can actually refer to the same physical visual contents and conceptions.

Unsupervised machine translation aims to learn cross-lingual mapping without the use of large-scale parallel corpora.
The setting is practically meaningful to those minor languages with hard data accessibility.
The basic idea is to leverage alternative pivoting contents to compensate the parallel signals based on the back-translation method \cite{sennrich-etal-2016-improving}, such as third-languages \cite{li-etal-2020-reference}, bilingual lexicons \cite{LampleCDR18} or multilingual LM \cite{ConneauL19}.
The visual information can also serve as pivot signals for UMT, i.e., unsupervised multimodal machine translation.
Comparing to the standard MMT that trains with <\emph{src-img-tgt}> triples, UMMT takes as input only the <\emph{src-img}>.
So far, few studies have explored the UMMT setting, most of which try to enhance the back-translation with multimodal alignment mechanism \cite{NakayamaN17,ChenLL18,SuFBKH19,huang-etal-2020}.

Scene graph describes a scene of an image or text into a structure layout, by connecting discrete objects with attributes and with other objects via pairwise relations \cite{KrishnaZGJHKCKL17,wang-etal-2018-scene}.
As the SGs carry rich contextual and semantic information, they are widely integrated into downstream tasks for enhancements, e.g., image retrieval \cite{JohnsonKSLSBL15}, image generation \cite{JohnsonGF18} and image captioning \cite{YangTZC19}.
This work inherits wisdom, incorporating both the visual scene graph and language scene graph as pivots for UMMT.

All the UMMT researches assume that the <\emph{src-img}> pairs are required during inference, yet we notice that this can be actually unrealistic.
We thus propose a visual hallucination mechanism, achieving the inference-time image-free goal.
There are relevant studies on supervised MMT that manage to avoid image inputs (with text only) during inference.
The visual retrieval-base methods \cite{0001C0USLZ20,fang-feng-2022-neural}, which maintain an image lookup-table in advance, such that a text can retrieve the corresponding visual source from the lookup-table.
\citet{LiPKCFCV22} directly build pseudo image representations from the input sentence.
Differently, we consider generating the visual scene graph with richer and holistic visual structural information.

\vspace{-1mm}
\section{Scene Graph-based Translation System}

\vspace{-1mm}
\subsection{Problem Definition}

In UMMT, no parallel translation pairs are available.
This work considers an inference-time image-free UMMT.
During training, the data availability is <$x,z$>$\in$<$\mathcal{X,Z}$> and the corresponding src-side LSG$^x$ and VSG, where $\mathcal{X}$ are the src-side sentences, and $\mathcal{Z}$ are the paired images.
During inference, the model generates tgt-side sentences $y \in\mathcal{Y}$ based on the inputs of only $x \in\mathcal{X}$ and the corresponding LSG$^x$, while the visual scene VSG$^{'}$ is hallucinated from LSG$^x$.
In both training and inference, $y$ will be generated from the intermediate tgt-side language scene graph LSG$^y$, which is produced from LSG$^x$ and VSG (or VSG$^{'}$).

\vspace{-1mm}
\subsection{Framework}
\label{sec:Framework}

\vspace{-1mm}
As shown in Fig. \ref{framework}, the system 
first represents the src-side LSG$^x$ and VSG features with two GCN graph encoders, respectively.
Then the SG fusing\&mapping module integrates and transforms two SG representations into a unified one as tgt-side LSG, i.e., LSG$^y$.
Another GSN model further encodes the LSG$^y$, where the representations are used to generate the tgt sentence (i.e., translation).

\begin{figure}[!t]
\centering
\includegraphics[width=0.98\columnwidth]{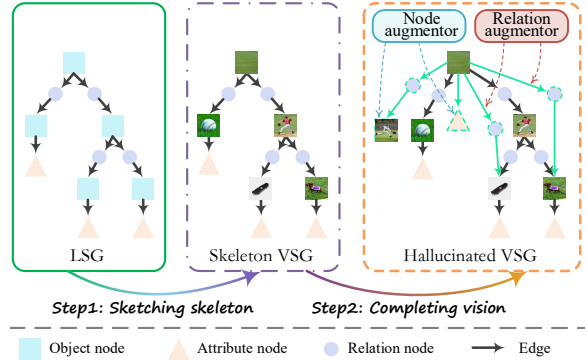}
\vspace{-1mm}
\caption{
The illustration of the visual scene hallucination (VSH) module, including two steps of inference.
}
\label{Visual-Scene-Hallucination}
\vspace{-3mm}
\end{figure}

\begin{figure*}[!t]
\centering
\includegraphics[width=0.98\textwidth]{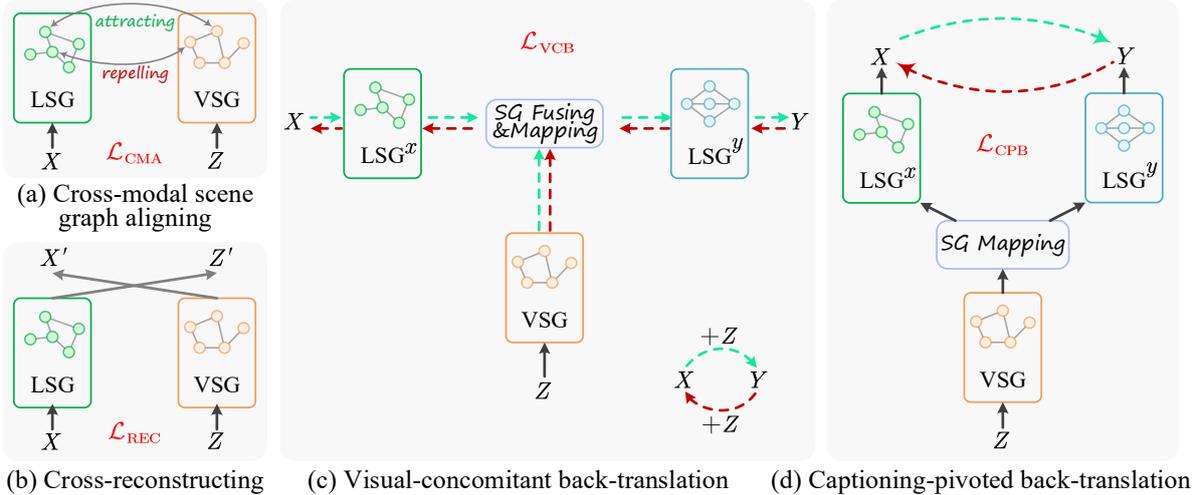}
\vspace{-1mm}
\caption{
Illustrations of the learning strategies for unsupervised multimodal machine translation.
}
\label{learning}
\vspace{-3mm}
\end{figure*}

\vspace{-1mm}
\paragraph{Scene Graph Generating and Encoding}
We first employ two off-the-shelf SG parsers to obtain the LSG and VSG, separately (detailed in the experiment part).
For simplicity, 
here we unify the notations of LSG and VSG as SG.
We denote a SG as $G$=($V,E$), where $V$ are the nodes (including object $o$, attribute $a$ and relation $r$ types), and $E$ are the edges $e_{i,j}$ between any pair of nodes $v_i \in V$.

We then encode both the VSG and LSG with two spatial Graph Convolution Networks (GCN) \cite{marcheggiani-titov-2017-encoding} respectively, which is formulated as:
\begin{equation}\label{SGCN}
\setlength\abovedisplayskip{3pt}
\setlength\belowdisplayskip{3pt}
 \bm{r}_{1}, \cdots, \bm{r}_{n} = \text{GCN}( G ) \,,
\end{equation}
where $\bm{r}_{i}$ is the representation of node $v_i$.
We here denote $\bm{r}_{i}^L$ as LSG's node representation, and $\bm{r}_{i}^V$ as VSG's node representation.

\vspace{-1mm}
\paragraph{Visual Scene Hallucinating}
\label{Visual Scene Hallucinating}

During inference, the visual scene hallucination (VSH) module is activated to perform two-step inference to generate the hallucinated VSG$^{'}$, as illustrated in Fig. \ref{Visual-Scene-Hallucination}.

\textbf{\emph{Step1: sketching skeleton}}
aims to build the skeleton VSG.
We copy all the nodes from the raw LSG$^x$ to the target VSG, and transform the textual entity nodes into the visual object nodes.

\textbf{\emph{Step2: completing vision}}
aims to enrich and augment the skeleton VSG into a more realistic one.
It is indispensable to add new nodes and edges in the skeleton VSG,
since in real scenarios, visual scenes are much more concrete and vivid than textual scenes.
Specifically, we develop a node augmentor and a relation augmentor, where the former decides whether to attach a new node to an existing one, and the later decides whether to create an edge between two disjoint nodes.
To ensure the fidelity of the hallucinated VSG$^{'}$, during training, the node augmentor and relation augmentor will be updated (i.e., with the learning target $\mathcal{L}_{\text{\scriptsize{VSH}}}$) with the input LSG and VSG supervisions.
Appendix $\S$\ref{Extended VSH Module} details the VSH module.

\vspace{-1mm}
\paragraph{SG Fusing\&Mapping}
\label{SG Fusing-Mapping}

Now we fuse the heterogeneous LSG$^x$ and VSG into one unified scene graph with a mixed view.
The key idea is to merge the information from two SGs serving similar roles.
In particular, we first measure the representation similarity of each pair of <\emph{text-img}> nodes from two GCNs.
For those pairs with high alignment scores, we merge them as one by averaging their representations, and for those not, we take the union structures from two SGs.
This results in a pseudo tgt-side LSG$^y$.
We then use another GCN model for further representation propagation.
Finally, we employ a graph-to-text generator to transform the LSG$^y$ representations to the tgt sentence $y$.
Appendix $\S$\ref{SG Fusing-Mapping Module} presents all the technical details in this part.

\vspace{-2mm}
\section{Learning with Scene Graph Pivoting}
\label{Learning with Scene Graph Pivoting}

\vspace{-1mm}
In this part, based on the SG pivot we introduce several learning strategies to accomplish the unsupervised training of machine translation.
We mainly consider 1) cross-SG visual-language learning,
and 2) SG-pivoted back-translation training.
Fig. \ref{learning} illustrates these learning strategies.

\vspace{-2mm}
\subsection{Cross-SG Visual-language Learning}

The visual-language SG cross-learning aims to enhance the structural correspondence between the LSG and VSG.
Via cross-learning we also teach the SG encoders to automatically learn to highlight those shared visual-language information while deactivating those trivial substructures, i.e., denoising.

\vspace{-2mm}
\paragraph{Cross-modal SG Aligning}

The idea is to encourage the text and visual nodes that serve a similar role in VSG and LSG to be closer.
To align the fine-grained structures between SGs, we adopt the contrastive learning (CL) technique \cite{LogeswaranL18,yan-etal-2021-consert,FeiMatchStruICML22,huang-etal-2022-conversation}.
In particular, CL learns effective representation by pulling semantically close content pairs together, while pushing apart those different ones.
Technically, we measure the similarities between pairs of nodes from two VSG and LSG:
\setlength\abovedisplayskip{2pt}
\setlength\belowdisplayskip{2pt}
\begin{equation}\small\label{similar-scoring-1}
s_{i,j} = \frac{ (\bm{r}^{L}_i)^T \cdot \bm{r}^{V}_j }{ || \bm{r}^{L}_i ||  \, || \bm{r}^{V}_j ||  }  \,.
\end{equation}
A threshold value $\alpha$ is pre-defined to decide the alignment confidence, i.e., pairs with $s_{i,j}>\alpha$ are considered similar.
Then we put on the CL loss:
\setlength\abovedisplayskip{2pt}
\setlength\belowdisplayskip{2pt}
\begin{equation}\small\label{CL-1}
\mathcal{L}_{\text{\scriptsize CMA}}  = - \sum_{i\in \text{LSG}^{x} ,\, j^{\ast}\in \text{VSG}} \log 
\frac{\exp(s_{i,j^{\ast}} /\tau )}{\mathcal{Z}}  \,,
\end{equation}
\begin{equation}\small\label{CL-2}
\mathcal{Z} = \sum_{i\in \text{LSG}^{x} ,\, k\in \text{VSG},\, k \ne j^{\ast}}\exp(s_{i,k}/\tau)  \,,
\end{equation}
where $\tau$>0 is an annealing factor.
$j^{\ast}$ means a positive pair with $i$, i.e., $s_{i,j^{\ast}}$>$\alpha$.

\vspace{-2mm}
\paragraph{Cross-modal Cross-reconstruction}

We further strengthen the correspondence between VSG and LSG via cross-modal cross-reconstruction.
Specifically, we try to reconstruct the input sentence from the VSG, and the image representations from the LSG.
In this way we force both two SGs to focus on the VL-shared parts.
To realize VSG$\to$$x$ we employ the aforementioned graph-to-text generator.
For LSG$\to$$z$, we use the graph-to-image generator \cite{JohnsonGF18}.
The learning loss can be marked as $\mathcal{L}_{\text{\scriptsize REC}}$.

\vspace{-2mm}
\subsection{SG-pivoted Back-translation Training}

\vspace{-1mm}
Back-translation is a key method to realize unsupervised machine translation \cite{sennrich-etal-2016-improving}.
In this work, we further aid the back-translation with structural SG pivoting.

\vspace{-2mm}
\paragraph{Visual-concomitant Back-translation}

We perform the back-translation with the SG pivoting.
We denote the $\mathcal{X}$$\to$$\mathcal{Y}$ translation direction as ${y}$=$\mathcal{F}^{xz\to y}$($x,z$), and $\mathcal{Y}$$\to$$\mathcal{Z}$ as ${x}$=$\mathcal{F}^{yz\to x}$($y,z$).
As we only have src-side sentences, the back-translation is uni-directional, i.e., $x$$\to$$\bar{y}$$\to$$\overline{\overline{x}}$.
\setlength\abovedisplayskip{2pt}
\setlength\belowdisplayskip{2pt}
\begin{equation}
\mathcal{L}_{\text{\scriptsize VCB}} = \mathbb{E} [-\log  p^{yz\to x} (\overline{\overline{x}} | \mathcal{F}^{xz\to y}(x,z) , z) ] \,.
\end{equation}

\vspace{-2mm}
\paragraph{Captioning-pivoted Back-translation}

Image captioning is partially similar to MMT besides the non-text part of the input.
Inspired by \citet{huang-etal-2020}, 
based on the SG pivoting, we incorporate two captioning procedures, $\mathcal{Z}$$\to$$\mathcal{X}$ and $\mathcal{Z}$$\to$$\mathcal{Y}$, to generate pseudo parallel sentences <$\bar{x}$-$\bar{y}$> for back-translation and better align the language latent spaces.
We denote $\mathcal{Z}$$\to$$\mathcal{X}$ as $\bar{x}$=$\mathcal{C}^{z\to x}$($z$), $\mathcal{Z}$$\to$$\mathcal{Y}$ as $\bar{y}$=$\mathcal{C}^{z\to y}$($z$).
The back-translation loss will be:
\setlength\abovedisplayskip{2pt}
\setlength\belowdisplayskip{2pt}
\begin{equation} \label{CPB}
\begin{aligned}
\mathcal{L}_{\text{\scriptsize CPB}} &= \mathbb{E} [-\log  p(\bar{x} | \mathcal{F}^{xz\to y}(\bar{x},z) , z) ] \, \\
 &+ \mathbb{E} [-\log  p( \bar{y} | \mathcal{F}^{yz\to x}(\bar{y} , z) , z) ] \,.
\end{aligned}
\end{equation}

\vspace{-2mm}
\paragraph{$\bigstar$ Remarks}

In the initial stage, each of the above learning objectives will be executed separately, in a certain order, so as to maintain a stable and effective UMMT system.
We first perform $\mathcal{L}_{\text{\scriptsize CMA}}$ and $\mathcal{L}_{\text{\scriptsize REC}}$, because the cross-SG visual-language learning is responsible for aligning the VL SGs, based on which the high-level translation can happen.
Then we perform back-translation training $\mathcal{L}_{\text{\scriptsize VCB}}$ and $\mathcal{L}_{\text{\scriptsize CPB}}$, together with VSH updating $\mathcal{L}_{\text{\scriptsize{VSH}}}$.
Once the system tends to converge, we put them all together for further fine-tuning:
\begin{equation}
\mathcal{L} = \mathcal{L}_{\text{\scriptsize CMA}} + \mathcal{L}_{\text{\scriptsize REC}} + \mathcal{L}_{\text{\scriptsize VCB}} + \mathcal{L}_{\text{\scriptsize CPB}} + \mathcal{L}_{\text{\scriptsize VSH}}\,.
\end{equation}

\section{Experiments}
\label{Experiments}

\vspace{-1mm}
\subsection{Setups}

\vspace{-1mm}
The experiments are carried out on Multi30K data \cite{elliott-etal-2016-multi30k}, a benchmark for MMT, where each image comes with three parallel descriptions in English/German/French.
Following \citet{huang-etal-2020}, we mainly consider the English-French (En$\leftrightarrow$Fr) and English-German (En$\leftrightarrow$De).
For each translation direction, we only use the src sentence \& img for training, and only the src sentence for testing.
We also test on the WMT16 En$\to$Ro and WMT14 En$\to$De, En$\to$Fr.
WMT \cite{bojar-etal-2014-findings,bojar-etal-2016-findings} is widely-used text-only translation corpora, where following \citet{LiPKCFCV22}, we use CLIP \cite{RadfordKHRGASAM21} to retrieve images from Multi30K for sentences.

Following prior research, we employ the FasterRCNN \cite{RenHGS15} as an object detector, and MOTIFS \cite{ZellersYTC18} as a relation classifier and an attribute classifier,
where these three together form a VSG generator.
For LSG generation, we convert the sentences into dependency trees with a parser \cite{00010BT0GZ18}, which is then transformed into the scene graph based on certain rules \cite{schuster-etal-2015-generating}.
For text preprocessing, we use Moses \cite{koehn-etal-2007-moses} for tokenization and apply the byte pair encoding (BPE) technique.
We use Transformer \cite{VaswaniSPUJGKP17} as the underlying text-encoder to offer representations for GCN, and use the FasterRCNN to encode visual feature representations. 
All GCN encoders and other feature embeddings have the same dimension of 1,024, and all GCN encoders are with two layers.

\begin{table*}[!t]
  \centering
\fontsize{9}{10.5}\selectfont
\setlength{\tabcolsep}{2.4mm}
\resizebox{0.96\textwidth}{!}{
\begin{tabular}{lcccccccc}
\hline
\multicolumn{1}{c}{\multirow{2}{*}{\textbf{}}}& \multicolumn{2}{c}{\textbf{En $\to$ Fr}} & \multicolumn{2}{c}{\textbf{En $\gets$ Fr}} & \multicolumn{2}{c}{\textbf{En $\to$ De}} & \multicolumn{2}{c}{\textbf{En $\gets$ De}} \\
\cmidrule(r){2-3}\cmidrule(r){4-5}\cmidrule(r){6-7}\cmidrule(r){8-9}
 & BLEU	& METEOR  & BLEU	& METEOR  & BLEU	& METEOR  & BLEU	& METEOR  \\
\hline
\multicolumn{9}{l}{$\bullet$ \textbf{\emph{Testing with image input given}}} \\
Game-MMT & 	- & 	- & 	- & 	- & 	16.6 & 	- & 	19.6 & 	- \\
UMMT & 	39.8 & 	35.5 & 	40.5 & 	37.2 & 	23.5 & 	26.1 & 	26.4 & 	29.7 \\
PVP & 	52.3  & 	67.6  & 	46.0  & 	39.8 &  	33.9  & 	54.1  & 	36.1 & 	34.7 \\
\rowcolor{nmgray}  \bf Ours$^\#$ & \bf 56.9 & 	\bf 70.7 & 	\bf 50.4 & 	\bf 42.5 & 	\bf 37.4 & 	\bf 57.2 & 	\bf 39.2 & 	\bf 38.3 \\
 \quad w/o SGs & 	 51.7 & 	64.0 & 	46.2 & 	40.7 & 	34.5 & 	56.4 & 	36.9 & 	35.2 \\
\hline

\multicolumn{9}{l}{$\bullet$ \textbf{\emph{Testing without image input given}}}\\
UMMT & 	15.8 & 	12.7 & 	10.2 & 	13.6 & 	8.4 & 	11.3 & 	7.5 & 	10.8 \\
UMMT$^*$ & 	30.4 & 	28.4 & 	31.8 & 	30.4 & 	15.7 & 	17.7 & 	19.3 & 	22.7 \\
PVP & 	26.1 & 	23.8 & 	25.7 & 	23.4 & 	11.1 & 	13.8 & 	14.0 & 	17.2 \\
PVP$^*$ & 	46.7 & 	58.0 & 	39.0 & 	31.9 & 	25.4 & 	40.1 & 	27.6 & 	26.0 \\
\rowcolor{nmgray}  \bf Ours & \bf 50.6 & \bf 64.7 & \bf 45.5 & 	\bf 37.3 & 	\bf 32.0 & 	\bf 52.3 & \bf 33.6 & \bf 32.8 \\
\specialrule{0em}{-2pt}{-2pt} & \scriptsize{(+3.9)}  & \scriptsize{(+6.7)}  & \scriptsize{(+6.5)} & \scriptsize{(+5.4)}  & \scriptsize{(+6.6)}  & \scriptsize{(+12.2)} & \scriptsize{(+6.0)}  & \scriptsize{(+6.8)}   \\
    \hline
    \end{tabular}%
    }
\vspace{-1mm}
    \caption{Results of UMMT on Multi30K data.
    `Ours$^\#$': using paired images for testing instead of visual hallucination.
    `UMMT$^*$/PVP$^*$': re-implemented baselines with phrase-level retrieval-based visual hallucination.
    In the brackets are the improvements of our model over the best-performing baseline(s).
    }
\vspace{-2mm}
  \label{tab:main}%
\end{table*}%

We mainly compare with the existing UMMT models: Game-MMT \cite{ChenLL18}, UMMT \cite{SuFBKH19} and PVP \cite{huang-etal-2020}.
To achieve a fair comparison on the inference-time image-free setup, we also re-implement the UMMT and PVP by integrating the phrase-level retrieval-based visual hallucination method \cite{fang-feng-2022-neural}.
All models use the same fair configurations, and we do not use pre-trained LM.
On WMT we also test the supervised MMT setup, where we use these baselines:
UVR \cite{0001C0USLZ20}, RMMT \cite{wu-etal-2021-good},  PUVR \cite{fang-feng-2022-neural} and VALHALLA \cite{LiPKCFCV22}.
We report the BLEU and METEOR scores for model evaluation.
Our results are computed with a model averaging over 5 latest checkpoints with significance test. 
Our experiments are based on the NVIDIA A100 Tensor Core GPUs.

\subsection{Main Results}

\begin{table}[t]
  \centering
\fontsize{9}{12.5}\selectfont
 \setlength{\tabcolsep}{0.6mm}
\resizebox{1\columnwidth}{!}{
    \begin{tabular}{lccccl}
    \hline
    \multicolumn{1}{c}{}& \multicolumn{1}{c}{\textbf{En$\to$Fr}} & \multicolumn{1}{c}{\textbf{En$\gets$Fr}} & \multicolumn{1}{c}{\textbf{En$\to$De}} & \multicolumn{1}{c}{\textbf{En$\gets$De}} & \multicolumn{1}{c}{\textbf{Avg.}} \\
    \hline

\rowcolor{nmgray} \bf Ours &	\bf 50.6 &	\bf 45.5 &	\bf 32.0 &	\bf 33.6 &	\bf 40.4 \\
- $L_{\text{\scriptsize CMA}}$ &	49.2  &	44.3  &	30.9  &	32.6  &	39.3\scriptsize{(-1.1)} \\
- $L_{\text{\scriptsize REC}}$ &	48.7  &	43.9  &	30.3  &	32.1  &	38.8\scriptsize{(-1.6)} \\
- $L_{\text{\scriptsize VCB}}$ &	47.0  &	42.2  &	28.7  &	30.1  &	37.0\scriptsize{(-3.4)} \\
- $L_{\text{\scriptsize CPB}}$ &	45.9  &	41.6  &	27.6  &	29.2  &	36.1\scriptsize{(-4.3)} \\
\cdashline{1-6}
- $L_{\text{\scriptsize CMA}}$\&$L_{\text{\scriptsize REC}}$ &	47.2  &	42.5 & 	29.2  &	30.9  &	37.5\scriptsize{(-2.9)} \\
- $L_{\text{\scriptsize CPB}}$\&$L_{\text{\scriptsize VCB}}$ &	44.6 &	40.0  &	26.3 & 	27.7 & 	34.7\scriptsize{(-5.7)} \\

 \hline
    \end{tabular}%
    }
    \caption{Ablating different learning strategies.
    }
    \vspace{-4mm}
  \label{tab:Ablation}%
\end{table}%

\paragraph{Results on Multi30K}

In Table \ref{tab:main} we show the overall results on Multi30K data.
First, we inspect the performance where gold-paired images are given as inputs for testing.
We see that our method (\emph{Ours$^\#$}), by integrating the LSG and VSG information, shows clear superiority over baselines on all translation jobs, while ablating the SGs, the performance drops rapidly.
This shows the importance of leveraging scene graphs for more effective multimodal feature representations.
Then, we look at the results where no paired images are given, i.e., an inference-time image-free setup.
By comparing \emph{UMMT/PVP} with \emph{UMMT$^*$/PVP$^*$} we understand that without images unsupervised MMT fails dramatically.
Notably, our system shows significant improvements over the best baseline \emph{PVP$^*$}, by average 5.75=(3.9+6.5+6.6+6.0)/4 BLEU score.
Although \emph{UMMT$^*$} and \emph{PVP$^*$} acquire visual signals via the phrase-level retrieval technique, our SG-based visual hallucination method succeeds much more prominently.
Besides, there are comparably small gaps between \emph{Ours} and \emph{Ours$^\#$}, which indicates that the proposed SG-based visual hallucination is highly effective.
The above observations prove the efficacy of our overall system for UMMT.

\begin{table}[!t]
  \centering
\fontsize{9}{11.5}\selectfont
 \setlength{\tabcolsep}{1.3mm}
\resizebox{0.95\columnwidth}{!}{
    \begin{tabular}{lcccc}
    \hline
    \multicolumn{1}{c}{}& \multicolumn{1}{c}{\textbf{En$\to$Ro}} &  \multicolumn{1}{c}{\textbf{En$\to$De}} & \multicolumn{1}{c}{\textbf{En$\to$Fr}} & \multicolumn{1}{c}{\textbf{Avg.}} \\
    \hline
\multicolumn{5}{l}{$\bullet$ \textbf{\emph{Supervised training (with parallel sentences)}}}\\

UVR &	\bf 33.8 &	28.2 &	39.6 &	33.8 \\
RMMT &	- &	24.5 &	35.3 &	- \\
PUVR &	33.2 &	\bf 28.5 &	39.9 &	\bf 33.9 \\
VALHALLA &	- &	28.0 &	\bf 40.0 &	- \\

\hline
\multicolumn{5}{l}{$\bullet$ \textbf{\emph{Unsupervised training (without parallel sentences)}}}\\

UMMT$^*$ &	27.4 &	20.8 &	32.6 &	26.9 \\
PVP$^*$ &	29.9 &	23.4 &	35.0 &	29.4 \\
\rowcolor{nmgray} \bf Ours & \bf 33.1	& \bf 27.8	& \bf 38.1	& \bf 33.0 \\

 \hline
    \end{tabular}%
    }
\vspace{-1mm}
    \caption{
    Results (BLEU) on WMT datasets.
    All model \textbf{supports inference-time image-free setting} with visual hallucination mechanism.
    }
    \vspace{-3mm}
  \label{tab:WMT}%
\end{table}%

\begin{figure*}[!t]
\centering
\includegraphics[width=0.98\textwidth]{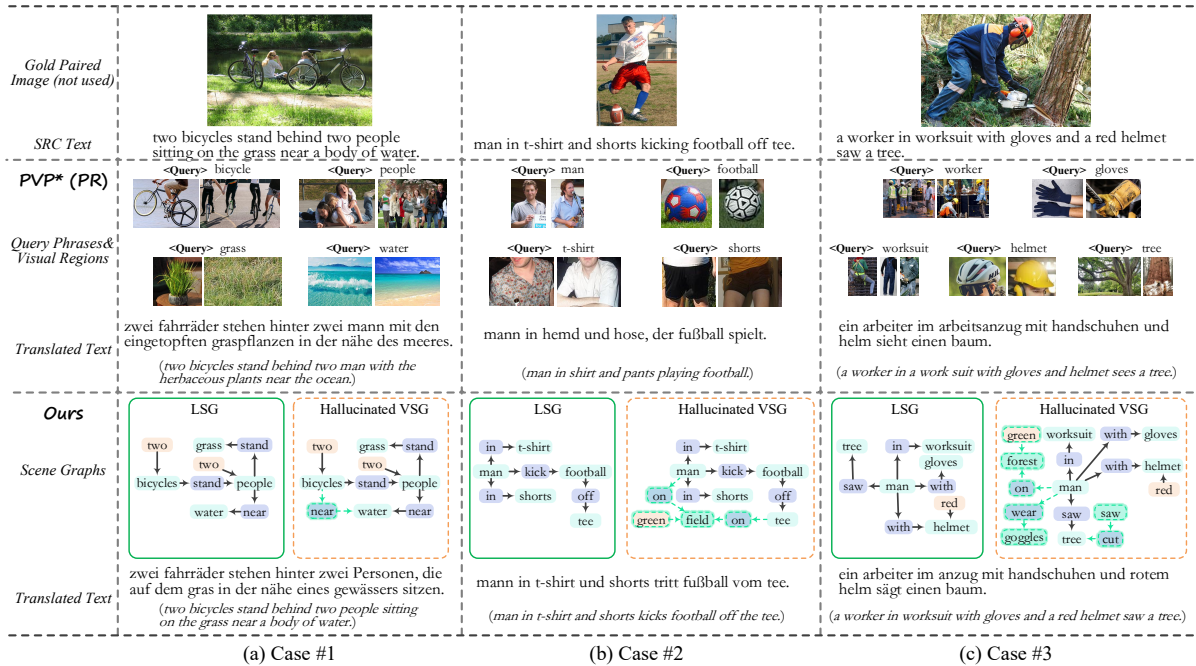}
\vspace{-1mm}
\caption{
Qualitative results of inference-time image-free UMMT (En$\to$De).
}
\label{case-study}
\vspace{-2mm}
\end{figure*}

\paragraph{Ablation Study}
In Table \ref{tab:Ablation} we quantify the contribution of each objective of scene graph pivoting learning via ablation study.
Each learning strategy exhibits considerable impacts on the overall performance, where the captioning-pivoted back-translation influences the results the biggest, with an average 4.3 BLEU score.
Overall, two SG-pivoted back-translation training targets show much higher influences than the two cross-SG visual-language learning objectives.
When removing both two back-translation targets, we witness the most dramatic decrease, i.e., average -5.7 BLEU.
This validates the long-standing finding that the back-translation mechanism is key to unsupervised translation \cite{sennrich-etal-2016-improving,huang-etal-2020}.

\begin{table}[!t]
  \centering
\fontsize{9}{12}\selectfont
\setlength{\tabcolsep}{1.0mm}
\resizebox{1\columnwidth}{!}{
\begin{tabular}{lcccc}
\hline
& \multicolumn{1}{c}{\multirow{2}{*}{\tabincell{c}{\bf Avg.\\BLEU}}}& \multicolumn{3}{c}{\textbf{Human evaluation}} \\
\cline{3-5}
 &  & \bf Completeness$\uparrow$  & \bf Ambiguity$\downarrow$	& \bf Fluency$\uparrow$ \\
\hline
PVP$^*$(SR) & 	33.2 & 	7.1 & 	7.6 & 	8.0 \\
PVP$^*$(PR) & 	35.0 & 	7.8 & 	5.0 & 	8.5 \\
\rowcolor{nmgray}  \bf Ours & 	\bf 39.3 & 	\bf 9.2$^{\dagger}$ & \bf 2.5$^{\dagger}$ & \bf 9.7$^{\dagger}$ \\
\quad w/o SG & 	35.7 & 	7.6 & 	6.7 & 	8.6 \\
\hline
\end{tabular}%
}
\caption{
Human evaluations are rated on a Likert 10-scale, where the results are averaged on En$\to$De and De$\to$En.
PVP$^*$ model uses the \underline{s}entence-level and \underline{p}hrase-level \underline{r}etrieval-based visual hallucination (i.e., SR and PR), respectively, during testing.
$\dagger$ indicates significance over the variants.
}
\vspace{-3mm}
  \label{human}%
\end{table}%

\paragraph{Results on WMT}
Table \ref{tab:WMT} further compares the translation results on WMT corpora under supervised/unsupervised MMT.
It is unsurprising to see that MMT models trained with supervision from parallel sentences are overall better than the unsupervised ones.
However, our UMMT system effectively narrows the gap between supervised and unsupervised MMT.
We can find that our unsupervised method only loses within 1 BLEU score to supervised models, e.g., \emph{UVR} and \emph{PUVR}.

\subsection{Further Analyses and Discussions}

In this part we try to dive deep into the model, presenting in-depth analyses to reveal what and how our proposed method really works and improves.

\begin{table}[!t]
  \centering
\fontsize{9}{12.5}\selectfont
 \setlength{\tabcolsep}{0.6mm}
\resizebox{1\columnwidth}{!}{
    \begin{tabular}{lcc}
    \hline
  & \bf Overall Txt-Img. & \bf Regional Phrase-Object\\
    \hline
PVP$^*$(SR) &	67.4\scriptsize{$\pm$6.8}  &	- \\
PVP$^*$(PR) &	-	 &88.9\scriptsize{$\pm$5.4} \\
\rowcolor{nmgray} \bf Ours & \bf 86.8\scriptsize{$\pm$4.7} &	\bf 91.4\scriptsize{$\pm$3.8} \\
- $L_{\text{\scriptsize CMA}}$ &	76.5\scriptsize{$\pm$5.5} &	80.3\scriptsize{$\pm$4.3} \\
- $L_{\text{\scriptsize REC}}$ &	70.1\scriptsize{$\pm$5.2} &	77.5\scriptsize{$\pm$4.0} \\
- $L_{\text{\scriptsize CMA}}$\&$L_{\text{\scriptsize REC}}$ &	68.6\scriptsize{$\pm$6.1} &	72.8\scriptsize{$\pm$4.8} \\

 \hline
    \end{tabular}%
    }
    \caption{
Vision-language aligning evaluation.
For our models, we transform the hallucinated VSG into an image via a graph-to-image generator.
We use CLIP to measure the VL relevance score.
    }
\vspace{-3mm}
  \label{align}%
\end{table}%

\vspace{-1mm}
\paragraph{$\bullet$ Integration of the vision and language SGs helps gain a holistic understanding of input.}
Both VSG and LSG advance in comprehensively depicting the intrinsic structure of the content semantics, which ensures a holistic understanding of the input texts and images.
By encoding the vision and language SGs, it is expected to completely capture the key components from src inputs, and thus achieve better translations.
However, without such structural features, some information may be lost during the translation.
In Table \ref{human} via human evaluation we can see that our system obtains significantly higher scores in terms of the \emph{completeness}, comparing to those baselines without considering SGs.
Also in Fig. \ref{case-study}, we can find that the baseline system \emph{PVP$^*$(PR)}, with only the local-level phrase-level visual retrieval, has frequently missed the key entities during the translation, e.g., the object `\texttt{tee}' in case\#2.

\vspace{-1mm}
\paragraph{$\bullet$ SG-based multimodal feature modeling helps achieve more accurate alignment between vision and language.}
Another merit to integrating the SGs is that the fine-grained graph modeling of visual and language scenes obviously aids more precise multimodal feature alignment.
In this way, the translated texts have higher fidelity to the original texts.
Inaccurate multimodal alignment without considering the SG modeling will otherwise lead to worse ambiguity.
Observing the \emph{ambiguity} in Table \ref{human}, we see that our model exhibits the lowest ambiguity.
In Fig. \ref{case-study} for the case\#3, \emph{PVP$^*$(PR)} confuses the verb `\texttt{saw}' as `\texttt{see}' as it fails to accurately refer `\texttt{saw}' to \emph{a certain lumbering tool}, while ours gives a correct prediction.
Besides, accurate multimodal alignment greatly enhances the utility of visual information. 
In Table \ref{align} we compare the relevance of vision-language counterparts by different models, where our model gives the highest performance on both the overall text-image matching and the regional phrase-object matching.
In addition, two proposed cross-SG learning targets display big impacts on the VL-aligning ability.

\begin{figure}[!t]
\centering
\includegraphics[width=0.97\columnwidth]{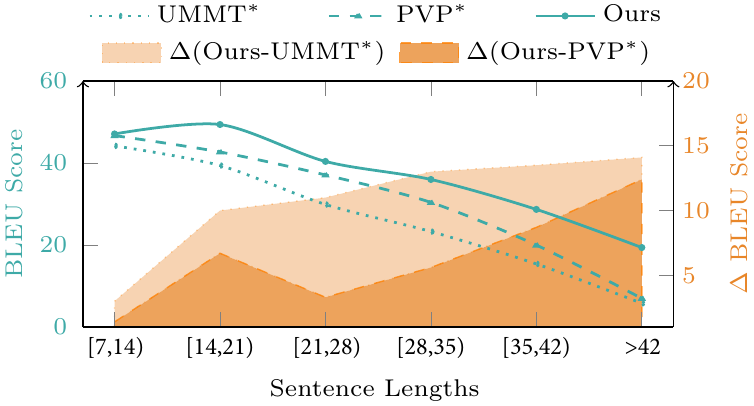}
\vspace{-1mm}
\caption{
BLEU scores under different sentence lengths.
}
\label{sent-length}
\vspace{-3mm}
\end{figure}

\vspace{-1mm}
\paragraph{$\bullet$ The longer and more complex the sentences, the higher the translation quality benefiting from the SGs features.}
In this work, we investigate the SG structures to model the input texts.
Graph modeling of the texts has proven effective for resolving the long-range dependency issue \cite{marcheggiani-titov-2017-encoding,LiPKCFCV22}.
In Fig. \ref{sent-length} we group the translation performance based on the lengths of source sentences.
We see that our SG-based model gives very considerable gains over the two non-SG baselines, where the longer the sentences the higher the improvements.

\vspace{-1mm}
\paragraph{$\bullet$ Incorporating SGs into MMT advances in more fluent translation.}
Also, modeling the semantic scene graph of the input features contributes a lot to the language fluency of the translation texts.
Looking at the \emph{Fluency} item in Table \ref{human}, we find that our system gives the best fluency with the lowest grammar errors.

\begin{figure}[!t]
\centering
\includegraphics[width=0.97\columnwidth]{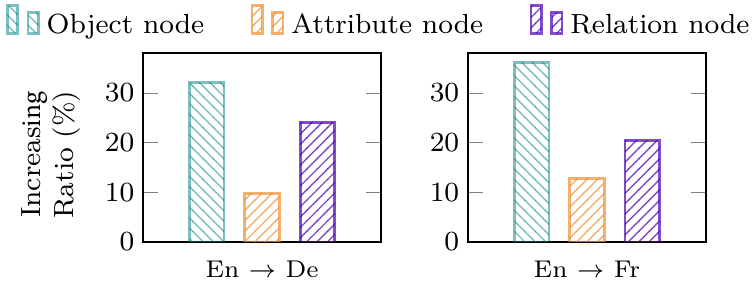}
\vspace{-1mm}
\caption{
Growing rate of nodes in hallucinated VSG.
}
\label{node-adding}
\vspace{-1mm}
\end{figure}

\begin{figure}[!t]
\centering
\includegraphics[width=0.87\columnwidth]{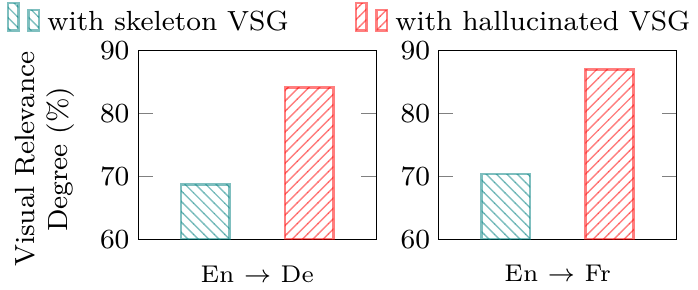}
\vspace{-1mm}
\caption{
Degree of visual relevance (similarity) between the hallucinated vision (via graph-to-image generator) and the ground truth image.
}
\label{reduc-degree}
\vspace{-1mm}
\end{figure}

\vspace{-1mm}
\paragraph{$\bullet$ SG-based visual scene hallucination mechanism helps gain rich and correct visual features.}
Different from the baseline retrieval-based methods that directly obtain the whole images (or local regions), our proposed VSH mechanism instead compensatively generates the VSGs from the given LSGs.
In this way, the hallucinated visual features enjoy two-fold advantages.
On the one hand, the pseudo VSG has high correspondence with the textual one, both of which will enhance the shared feature learning between the two modalities.
On the other hand, the hallucinated VSG will produce some vision-specific scene components and structures, providing additional clues to facilitate back to the textual features for overall better semantic understanding.
Fig. \ref{node-adding} illustrates the node increasing rate during the vision scene graph hallucination.
We see that the numbers of all three types of nodes increase, to different extents, where object nodes grow rapidest.
Also, during the two transition steps of the VSH mechanism we get two VSGs, skeleton VSG and hallucinated VSG.
From Fig. \ref{reduc-degree} we see that after two full hallucination steps, we can obtain high-fidelity vision features, demonstrating the necessity of the second \emph{completing-vision} step.

\vspace{-1mm}
\section{Conclusion}

\vspace{-1mm}
We investigate an \emph{inference-time image-free} setup in unsupervised multimodal machine translation.
In specific, we integrate the visual and language scene graph to learn the fine-grained vision-language representations.
Moreover, we present a visual scene hallucination mechanism to generate pseudo visual features during inference.
We then propose several SG-pivoting learning objectives for unsupervised translation training. 
Experiments demonstrate the effectiveness of our SG-pivoting based UMMT.
Further experimental analyses present a deep understanding of 
how our method advances the task and setup.

\section*{Acknowledgments}

This research is supported by the National Natural Science Foundation of China (No. 62176180), and also the Sea-NExT Joint Lab.

\section*{Limitations}

Our paper has the following potential limitations.
First of all, we take advantage of the external scene graph structures to achieve the inference-time visual hallucination and secure significant improvements of the target task, while it could be a double-edged sword.
This makes our method subject to the quality of the external structure parsers.
When the parsed structures of visual scene graphs and language scene graphs are with much noise, it will deteriorate our methods.
Fortunately, the existing scene graph parsers have already achieved satisfactory performance for the majority language (e.g., English), which can meet our demands.
Second, the effectiveness of our approach depends on the availability of good-quality images, which however shares the pitfalls associated with the standard unsupervised multimodal translation setup.

\bibliography{anthology}
\bibliographystyle{acl_natbib}

\newpage

\appendix

\section{Appendix}
\label{Model Details}

\begin{figure}[!t]
\centering
\includegraphics[width=1.0\columnwidth]{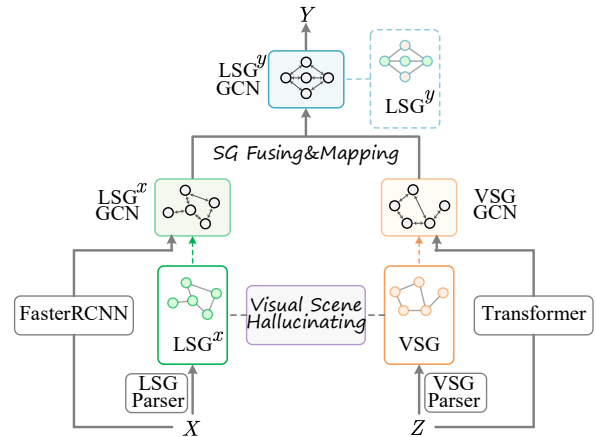}
\caption{
A detailed view of our model architecture.
The tgt-side LSG$^y$ is synthesized from input LSG$^x$ and VSG, which is a pseudo LSG without a real input of LSG$^y$ from a parser.
}
\label{app-all-framework}
\end{figure}

In $\S$\ref{sec:Framework} we give a brief induction to the overall model framework.
Here we extend the details of each module of the scene graph-based multimodal translation backbone.
In Fig. \ref{app-all-framework} we outline our framework.

\begin{figure*}[!t]
\centering
\includegraphics[width=0.78\textwidth]{illus/augmentor2.pdf}
\caption{
Illustrations of the node augmentor and the relation augmentor.
}
\label{augmentor}
\end{figure*}

\subsection{Visual Scene Hallucination Learning Module}
\label{Extended VSH Module}

First of all, we note that VSH only will be activated to produce VSG hallucination at inference time.
During the training phase, we construct the VSG vocabularies of different VSG nodes.
We denote the object vocabulary as $D^o$, which caches the object nodes from parsed VSG of training images;
denote the attribute vocabulary as $D^a$, which caches the attribute nodes;
and denote the relation vocabulary as $D^r$, which caches the relation nodes.
Those vocabularies will be used to provide basic ingredients for VSG hallucination.

At inference time, VSH is activated to perform two-step inference to generate the hallucinated VSG$^{'}$.
The process is illustrated in Fig. \ref{Visual-Scene-Hallucination}.

\paragraph{Step1: Sketching Skeleton}

This step builds the skeleton VSG from the raw LSG.
Specifically, we only need to transform the textual entity nodes into the visual object nodes, while keeping unchanged the whole graph topology.
As for the attribute nodes and the relation nodes, we directly copy them into the VSG, as they are all text-based labels that are applicable in VSG.
Then we transform the textual entity nodes into the visual object nodes.
For each textual entity node in LSG, we employ the CLIP tool\footnote{\url{https://github.com/openai/CLIP}} to search for the best matching visual node (proposal) in $D^o$ as the counterpart visual object, resulting in the skeleton VSG.
After this step, we obtain the sketch structure of the target VSG.

\paragraph{Step2: Completing Vision}

\begin{figure}[!t]
\centering
\includegraphics[width=1.0\columnwidth]{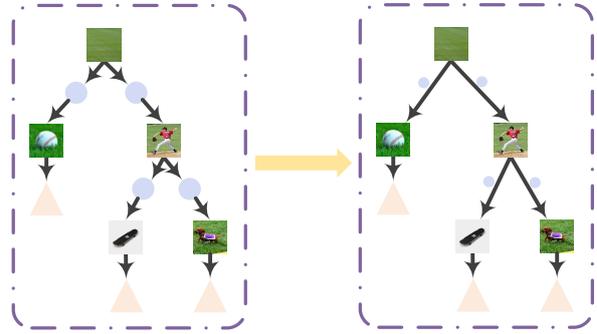}
\caption{
Degeneration of the relation node to the labeled edge.
}
\label{degeneration}
\vspace{-5mm}
\end{figure}

This step completes the skeleton VSG into a more realistic one, i.e., the final hallucinated VSG$^{'}$.
With the skeleton VSG at hand, we aim to further enrich skeleton VSG.
Because intuitively, in actual world the visual scenes are always much more concrete and vivid than textual scenes.
For example, given a caption text `\emph{boys are playing baseball on playground}', the LSG only mentions `\emph{boys}', `\emph{baseball}' and `\emph{playground}' objects.
But imaginarily, there must be a `\emph{baseball bat}' in the scene of vision, and also both the pairs of `\emph{boys}'-`\emph{playground}' and `\emph{baseball}'-`\emph{playground}' has `on' relation.
Thus it is indispensable to add new nodes and more edges, i.e.,  scene graph augmentation.
To reach the goal, we propose a \textbf{node augmentor} and a \textbf{relation augmentor}, as shown in Fig. \ref{augmentor}.
First of all, we downgrade all the relation nodes as the edge itself, i.e., an edge with a relation label.
By this, we obtain a VSG that only contains object and attribute nodes, and labeled edges, which is illustrated in Fig. \ref{degeneration}.

$\blacktriangleright$ For the node augmentor, we first traverse all the object nodes in the skeleton VSG.
For each object node $v_i$, we then perform \emph{$k$-order routing} over its neighbor nodes.
We denote its neighbor nodes as $V_i^{na}=\{\cdots,v_k,\cdots\}$.
Then we use the attention to learn the neighbor influence to $v_i$, and obtain the $k$-order feature representation $\bm{h}_i$ of $v_i$:
\begin{equation}
\begin{aligned} \nonumber
\alpha^n_k &= \frac{\exp{ \bm{r}_i \cdot \bm{r}_k }}{ \sum_{v_k^{*} \in V_i^{na}} \exp{ \bm{r}_i \cdot \bm{r}_k^{*} }  }   \, \\
\bm{h}^{na}_i &= \bm{r}_i + \sum_{k} \alpha^n_k \cdot \bm{r}_k \,.
\end{aligned}
\end{equation}
where $\bm{r}_i$ and $\bm{r}_k$ is the node representations of $v_i$ and $v_k$, which are obtained from GCN encoder.
Then we use a classifier to make prediction over the total vocabularies of $D^o$ and $D^a$, to determine which node $\hat{v}^{'}_i$ (either an object or an attribute node) should be attached to $v_i$, if any:
\begin{equation} \nonumber
\hat{v}^{'}_i \gets \underset{D^{na}}{\text{Softmax}}( \text{FFN} (\bm{h}^{na}_i) ) \,,
\end{equation}
where $D^{na}=D^o \cup D^a \cup \{\epsilon\}$, including an additional dummy token $\epsilon$ indicating no new node to be attached to $v_i$.
And if the predicted node is an object node, an additional relation classifier will determine what is the relation label $\hat{e}^{'}$ between $\hat{v}^{'}_i$ and $v_i$:
\begin{equation} \nonumber
\hat{e}^{'} \gets \underset{D^{r}}{\text{Softmax}}( \text{FFN} ([\bm{h}^{na}_i;\bm{r}_i) ) \,.
\end{equation}

$\blacktriangleright$ For the relation augmentor, we first traverse all the node-pairs (object or attribute nodes, excluding the relation nodes) in the VSG, i.e., $v_i \& v_j$.
Then, for each node in the pair we use a triaffine attention \cite{wang-etal-2019-second,Wu0RJL21} to directly determine which new relation type $\hat{e}^{'}_{i,j}$ should be built between them, if exists:
\begin{equation} \nonumber
\setlength\abovedisplayskip{2pt}
\setlength\belowdisplayskip{2pt}
\bm{h}^{pa}_{i-j} =  \text{Sigmoid} ( \left[
\begin{array}{c}
  \bm{r}_i    \\
    1
\end{array}
\right]^\mathrm{T}
(\bm{r}_j)^\mathrm{T}
\mathbf{W} \left[
\begin{array}{c}
  \bm{r}_{i-j}    \\
    1
\end{array}
\right]  )  \,,
\end{equation}
\begin{equation} \nonumber
\hat{e}^{'}_{i,j} \gets \underset{D^{pa}}{\text{Softmax}}( \text{FFN} (\bm{h}^{pa}_{i-j} ) ) \,,
\end{equation}
where $D^{pa}=D^r\cup\{\epsilon\}$, where the dummy token $\epsilon$ indicates no new edge should be created between two nodes.
The new edge $\hat{e}^{'}_{i,j}$ has a relation label.
$\bm{r}_{i-j}$ is the representation of the path from $v_i$ to $v_j$, which is obtained by the pooling function over all the nodes in the path: 
\begin{equation} \nonumber
\bm{h}^{pa}_{i-j} =  \text{Pool} (\bm{r}_i, \cdots,  \bm{r}_j ) \,.
\end{equation}
Note that the triaffine scorer is effective in modeling the high-order ternary relations, which will provide a precise determination on whether to add a new edge.

During training, the node augmentor and the relation augmentor are trained and updated based on the gold LSG and VSG, to learn the correct mapping between LSG and VSG.
\begin{equation}
\begin{aligned} \nonumber
\mathcal{L}_{\scriptsize{NA}} &= \sum [ \log p(\hat{v}^{'}_i| VSG\gets LSG  ) \\
&+ \log p(\hat{e}^{'}_{i,j}| VSG\gets LSG  ) ] \,,
\end{aligned}
\end{equation}
\begin{equation} \nonumber
\mathcal{L}_{\scriptsize{PA}} = \sum \log p(\hat{e}^{'}_{i,j} | VSG\gets LSG ) \,,
\end{equation}
\begin{equation} \nonumber
\mathcal{L}_{\text{\scriptsize{VSH}}} = \mathcal{L}_{\scriptsize{NA}} + \mathcal{L}_{\scriptsize{PA}} \,.
\end{equation}
Such supervised learning is also important for ensuring that the final generated hallucinated visual scenes are basically coincident with the caption text, instead of random or groundless vision scenes.

\subsection{SG Fusing\&Mapping Module}
\label{SG Fusing-Mapping Module}

Here we extend the contents in $\S$ \ref{SG Fusing-Mapping}.
As shown in Fig. \ref{app-all-framework}, first of all, the SG fusing module merges the LSG$^x$ and VSG into a mixed cross-modal scene graph, such that the merged scene graph are highly compact with less redundant.
Before the merging, we first measure the similarity of each pair of <\emph{text-img}> node representations via cosine distance:
\begin{equation}\nonumber
s^f_{i,j} = \frac{ (\bm{r}^{L}_i)^T \cdot \bm{r}^{V}_j }{ || \bm{r}^{L}_i ||  \, || \bm{r}^{V}_j ||  }  \,.
\end{equation}
This is a similar process as in Eq. (\ref{similar-scoring-1}).
For those pairs with high alignment scores, i.e., $s_{i,j}>\alpha$ (we use the same pre-defined threshold as in cross-modal alignment learning), we consider them as serving a similar role.
Since we will perform the cross-modal SG aligning learning $\mathcal{L}_{\text{\scriptsize CMA}}$, the accuracy of the alignment between LSG$^x$ and VSG can be guaranteed.
Then, we average the representations of the image-text node pair from their GCNs.
And for the rest of nodes in LSG$^x$ and VSG, we take the union structures of them.
The resulting mixed SG fully inherits the semantic-rich scene nodes from both the textual SG and the visual SG, which will benefit the following text generation.

Now we treat the mixed SG as a pseudo tgt-side LSG$^y$.
We use another GCN to model LSG$^y$ for further feature propagation:
\begin{equation}\nonumber
 \bm{r}^y_{1}, \cdots, \bm{r}^y_{m} = \text{GCN}( VSG^y ) \,.
\end{equation}
The initial node representations of GCN are from the GCNs of VSG and LSG$^x$, i.e., $\bm{r}^{L}$ and $\bm{r}^{V}$ as in Eq. (\ref{SGCN}).
Based on the node representation $\bm{r}^y_i$ of VSG$^y$, we finally employ a graph-to-text model\footnote{\url{https://github.com/freesunshine0316/neural-graph-to-seq-mp}} to generate the final tgt-side sentence.
Specifically, all the node representation $\bm{r}_i$ will be first summarized into one unified graph-level feature via pooling: 
\begin{equation} \nonumber
\bm{r}^{y} =  \text{Pool} ( \bm{r}_{1}^y, \cdots, \bm{r}^y_{m}) \,.
\end{equation}
Then, an autoregressive sequential decoder (SeqDec) will take $\bm{r}^{y}$ to generate tgt-side token over the tgt-side vocabulary at each setp, sequentially:
\begin{equation}\nonumber
\bm{e}_i = \text{SeqDec} (\bm{e}_{\le i} \,,\, \bm{r}^{y}) \,,
\end{equation}
\begin{equation}\nonumber
\hat{y}_{i} \gets \text{Softmax}(\bm{e}_i ) \,.
\end{equation}

\end{document}